\documentclass[conference]{IEEEtran}
\IEEEoverridecommandlockouts
\usepackage{cite}
\usepackage{amsmath,amssymb,amsfonts}
\usepackage{algorithmic}
\usepackage{graphicx}
\usepackage{textcomp}
\usepackage{xcolor}
\def\BibTeX{{\rm B\kern-.05em{\sc i\kern-.025em b}\kern-.08em
    T\kern-.1667em\lower.7ex\hbox{E}\kern-.125emX}}
\begin{document}

\title{GPT-4 Emulates Average-Human Emotional Cognition from a Third-Person Perspective \\
}

\author{\IEEEauthorblockN{Ala N. Tak and Jonathan Gratch}
\IEEEauthorblockA{\textit{Institute for Creative Technologies} \\
\textit{University of Southern California}\\
Playa Vista, CA 90094, USA\\
(antak, gratch)@ict.usc.edu}
}



\maketitle

\begin{abstract}
This paper extends recent investigations on the emotional reasoning abilities of Large Language Models (LLMs). Current research on LLMs has not directly evaluated the distinction between how LLMs predict the self-attribution of emotions and the perception of others' emotions. We first look at carefully crafted emotion-evoking stimuli, originally designed to find patterns of brain neural activity representing fine-grained inferred emotional attributions of others. We show that GPT-4 is especially accurate in reasoning about such stimuli. This suggests LLMs agree with humans' attributions of others' emotions in stereotypical scenarios remarkably more than self-attributions of emotions in idiosyncratic situations. To further explore this, our second study utilizes a dataset containing annotations from both the author and a third-person perspective. We find that GPT-4's interpretations align more closely with human judgments about the emotions of others than with self-assessments. Notably, conventional computational models of emotion primarily rely on self-reported ground truth as the gold standard. However, an average observer's standpoint, which LLMs appear to have adopted, might be more relevant for many downstream applications, at least in the absence of individual information and adequate safety considerations. 

\end{abstract}


\section{Introduction}
The exploration of large language models (LLMs) in understanding and modeling human emotions has received significant attention in the last two years. These studies have probed the capabilities of models such as the GPT family of LLMs and others in tasks related to causal reasoning \cite{binz}, emotional decision-making and appraisal theory \cite{ala1}, emotion classification\cite{nut1,broekens,nut2}, emotional intelligence \cite{wang}, emotional dialogue understanding \cite{zhao}, generation of emotional text \cite{gagne}, and more. A consistent method across these studies is the zero-shot approach (i.e., in-context learning) with prompt engineering, emphasizing the LLMs' ability to perform tasks without explicit training.

\begin{figure}[tb] 
\centerline{\includegraphics{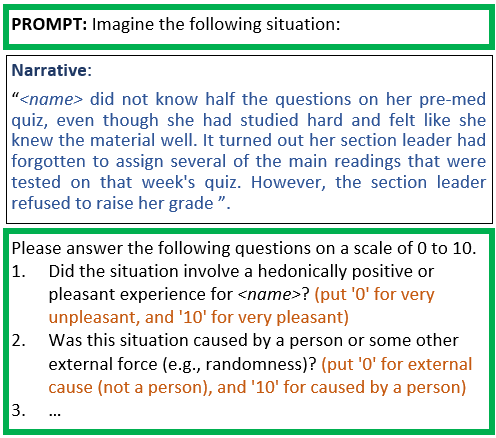}}
\caption{Example prompt}
\label{fig1}
\vspace{-10pt}
\end{figure}


\begin{figure*}[tb] 
\centerline{\includegraphics[width=1.95\columnwidth]{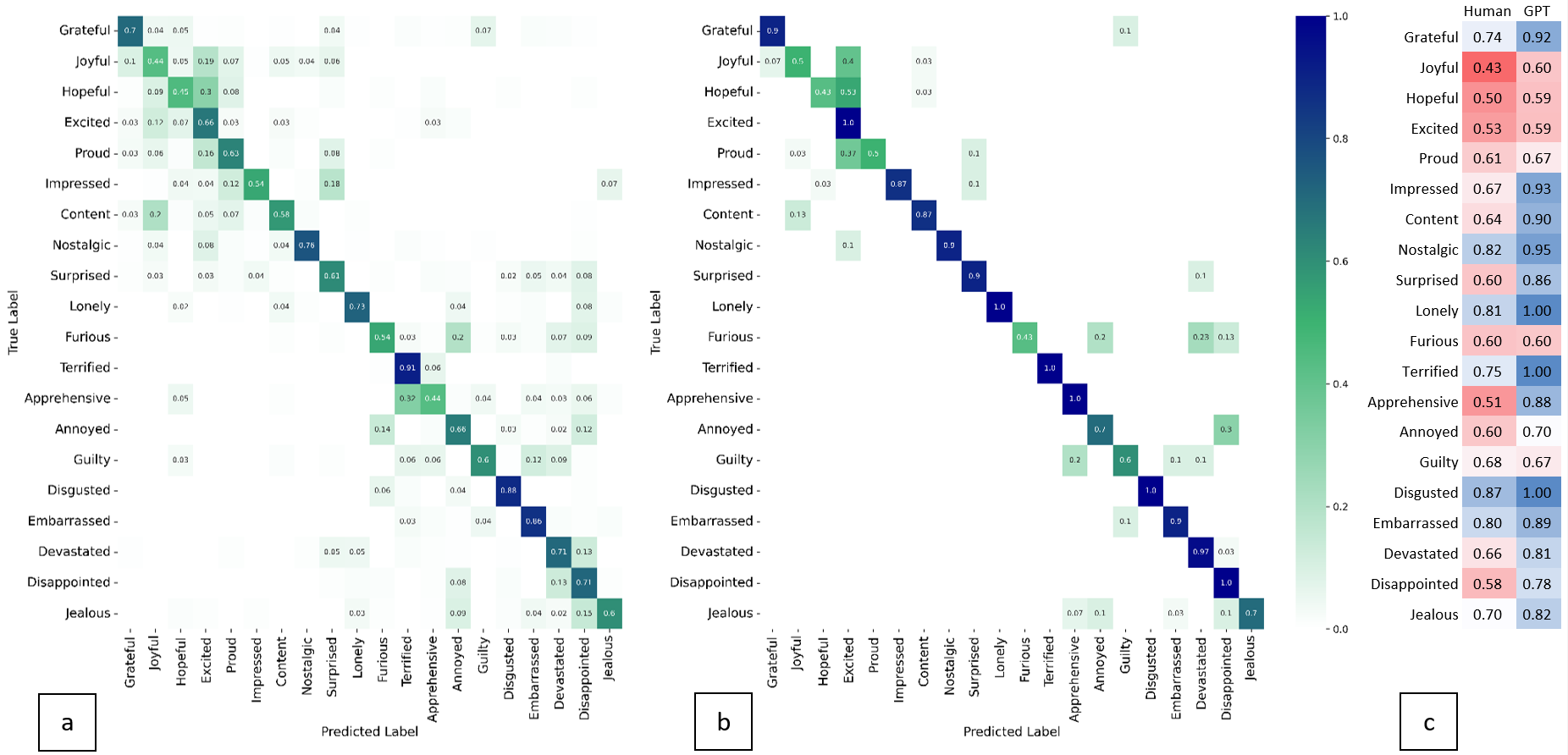}}
\caption{Comparison of human and GPT-4 stimuli classification; (a) human predicted labels, (b) GPT-4 predicted labels, (c) f-1 scores}
\label{confuse}
\vspace{-10pt}
\end{figure*}

While the field of affective computing often concentrates on inferring emotions from expressions, typically overlooking the triggering circumstances \cite{b1}, computational emotion models aim to understand the situational context, including how specific aspects may evoke particular emotions and influence future decisions, behaviors, and beliefs \cite{b2, b3}. The foundation of most computational emotion models is \textit{appraisal theory}\cite{b4, b5}, actually a cluster of theories that share the principle that emotions arise from an evaluation of how current circumstances impact the individual. This evaluation, based on criteria known as appraisal variables, assesses the relevance of a situation to one's goals, its alignment with these goals, and its predictability, among other factors. The specific pattern of these assessments gives rise to particular emotions. For instance, anger is triggered by goal-incongruent events when the person perceives control, whereas sadness emerges from a sense of powerlessness. The intensity of these emotions is further shaped by factors such as the importance of the threatened goal or the unexpectedness of the threat, leading to stronger emotional responses \cite{b6}.

In this paper, we seek to address a persistent controversy involving appraisal theory as to whether it reflects the actual mechanisms involved in human emotion elicitation \cite{lazarus}, or whether it serves as a folk psychological theory that observers use to interpret the emotions of others \cite{demelo, shlomo}, or if both perspectives are equally valid (echoing similar controversies as to whether emotion recognition methods are best seen as recognizing felt or perceived emotion). If the former, LLM-based models would be well-suited to emotion recognition. If the latter, they may be better suited to predicting social perceptions.

Current evaluations of LLM abilities have not directly evaluated the distinction between production and perception. For example, Tak and Gratch \cite{ala1, ala2} showcased the advantage of using appraisal theory as a lens to shed light on similarities and differences in how humans and LLMs attribute emotions to situations. Though the work only considered self-reported emotions from descriptions of autobiographical memories, and did not contrast the accuracy of these predictions against those of outside observers reading the same descriptions. This study was also limited by the small size of the corpus they used.

We address these limitations with two studies. We first look at carefully crafted, artificial emotion-evoking stimuli with ratings on a large set of appraisal and emotion dimensions. The stimuli were originally designed to find patterns of brain neural activity representing fine-grained inferred emotional attributions of others. We show that GPT-4 \cite{gpt4}, arguably the most capable LLM currently available, is more accurate in reasoning about such stimuli than free-form self-report vignettes. The performance might also be derived from the differences in perspective. In other words, GPT-4 might view situations as an observer and capture the third-person perspective of the average human. This hypothesis motivates the second study, in which we examine whether GPT-4 processes emotions through an average observer's lens. To this end, we employ a corpus that includes both author and reader annotations of appraisals and emotions. In both studies, we follow the current practice of assessing LLMs' zero-shot in-context learning abilities (temperature set to 0) employing OpenAI's API resources.

\begin{figure*}[tb] 
\vspace{-10pt}
\centerline{\includegraphics[width=1.6\columnwidth]{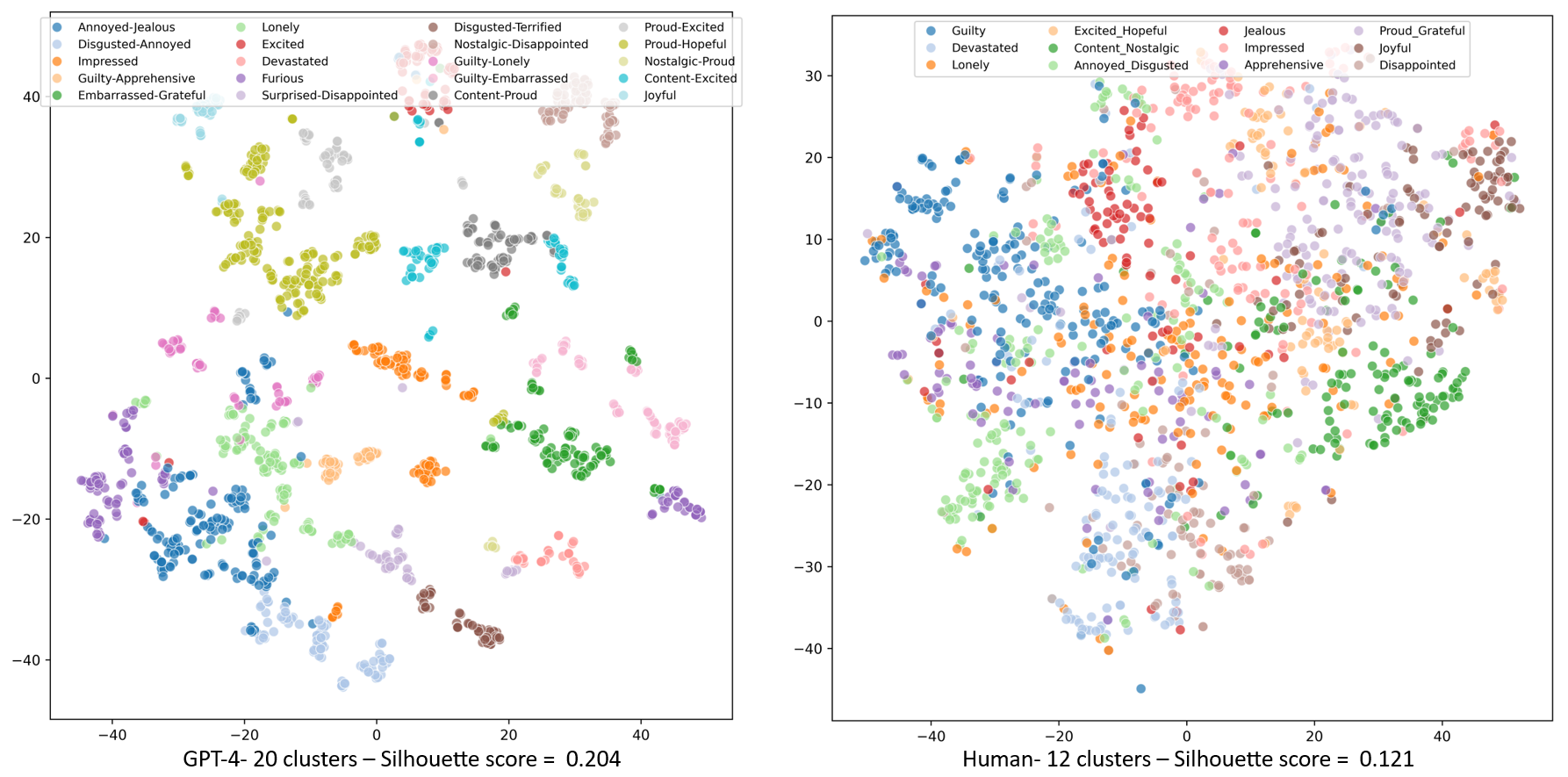}}
\caption{t-SNE plot generated using human and GPT-4 appraisal scores}
\label{tsne_comp}
\vspace{-10pt}
\end{figure*}

Our work builds upon recent efforts to unravel the underlying mechanisms and inner workings of LLMs and AI more broadly. Emotional inference plays a pivotal role across a spectrum of real-world social contexts, including courtroom judgments, therapeutic environments, negotiations, and personal relationships. Given its significance, the emotional cognitive capabilities of LLMs and AI at large can pose substantial risks or confer critical benefits. The perspective LLMs adopt in emotional inference is a fundamental component of their cognitive capacity, especially influencing their applicability in various domains. Depending on their intended use, an LLM might be optimally deployed either to recognize the emotions people are actually experiencing or to gauge social perceptions. For instance, a storytelling model should steer clear of endorsing any particular understanding of emotional experiences in the absence of a universal consensus, aiming instead to align with general social perceptions. Conversely, a model designed for personal therapy must closely align with an individual's authentic emotional state. The subsequent sections explore different components of LLM's emotional reasoning with discussions on LLM performance in relation to the nature of inputs (stereotypically crafted stimuli versus spontaneous, free-form scenarios) and the model's perspective (first-person/experiencer versus third-person/observer).

\section{Study 1: Crafted emotion-evoking stimuli}
Skerry and Saxe \cite{saxe} hypothesized that brain representations involved in inferring others' emotions based on short textual narratives are better captured by appraisal variables than by combinations of basic emotional dimensions. Instead of studying emotions as directly experienced by individuals, i.e., the authentic and subjective/first-person experience of emotions, Skerry and Saxe \cite{saxe} explored how people intuitively understand and theorize about the causes of emotions. In other words, they aimed to explore folk psychological theories\cite{folk} about emotions (i.e., how emotions are caused). Regardless of whether these ideas are directly tied to immediate emotional experiences, such folk psychological theories (e.g., appraisal theory) hold value as they often capture real causal regularities in the world.

In their study, subjects read 200 stimuli describing situations that would elicit a particular emotion. The reliability of the constructed stimuli was tested by a group of subjects on MTurk who classified the stimuli with 65\% accuracy (chance = 5\%). The constructed verbal stimuli (2-3 sentences each; M(SEM)= 50.68(0.28) words) describe a character in an emotion-eliciting event who experiences one of 20 different emotions without any mention of the character’s reaction. Participants (1521 total responses) are asked to rate the situation on high-dimensional appraisal space (38 variables drawn from different theories, particularly Scherer and colleagues \cite{scherer1}, \cite{scherer2}) and eight basic emotion dimensions (six basic emotions plus valence and arousal).

Aiming to replicate and extend the approach in \cite{ala1} on the extensive set of carefully crafted stimuli described above, we prompt GPT-4 to rate the scenarios using the same scales and wordings given to Skerry and Saxe's participants. We repeated each prompt eight times, yielding 1600 data points, enabling us to analyze variability in responses. An example prompt is illustrated in Fig.~\ref{fig1}, which includes the narrative each time generated with a random female name and a minimum additional text to help the model provide standardized output. Skerry and Saxe used random names to avoid bias that might arise from a particular name (like them, we did not examine any effects of name choice). In particular, we aim to examine how accurate GPT-4 would be at predicting people's assessment of others' appraisals (i.e., third-person appraisal derivation), people's attributions of others' emotion (i.e., third-person affect derivation), and would it be consistent with appraisal theory in explaining appraisal-emotion mapping?

\subsection{Reduced appraisal space}
Following Skerry and Saxe \cite{saxe}, we apply sequential feature selection to reduce appraisals to a smaller feature space as several of the 38 appraisals are highly correlated. 
The reduced appraisal space eliminates redundant features, 
helping to capture unique variance across stimuli.


Utilizing an ensemble classifier,
 we evaluate the contribution of each feature towards accurately classifying the 20 distinct emotion labels and incrementally add features that improve classification accuracy. A model trained on ten appraisal variables classifies the scenarios with 45.8\% accuracy compared to 56.6\% observed with the full appraisal space (thus, suggesting the reduced space achieves reasonable performance). Below is the list of selected features:

\begin{itemize}
\item \textbf{Pleasantness}: Did the situation involve a hedonically positive or pleasant experience for $\langle name\rangle$?
\item \textbf{Expectedness}: Did $\langle name\rangle$ expect this situation to occur?
\item \textbf{Agent-cause}: Was this situation caused by a person or some other external force (e.g., randomness)?
\item \textbf{Self-cause}: Was this situation caused by $\langle name\rangle$ herself or by someone/something else?
\item \textbf{Already-occurred}: Was $\langle name\rangle$'s emotion based on something that had already occurred?
\item \textbf{Close-others}: Did people other than $\langle name \rangle$ know about the situation that occurred?
\item \textbf{Pressure}: Was $\langle name\rangle$ under a lot of pressure in this situation?
\item \textbf{Consequences}: Was $\langle name\rangle$'s situation an isolated incident, or did it have long-term consequences?
\item \textbf{Safety}: Did this situation involve risks for $\langle name\rangle$ or others?
\item \textbf{Self-esteem}: Did this situation affect $\langle name\rangle$'s self-esteem or opinion of herself?
\end{itemize}

Strikingly, using the same ensemble approach with GPT-4 rated appraisals (i.e., using GPT-4 rather than humans to predict appraisal values)  achieved 94.5\% accuracy with the same reduced set, compared to 99.7\% accuracy using the full 39 appraisals. Fig.~\ref{confuse} illustrates the performance of GPT-4 the classify stimuli following their intended labels (i.e., True labels in the two confusion matrices). Fig.~\ref{confuse}.c denotes that GPT-4 outperforms human participants in attributing emotions to stimuli as they were intended by the researchers. It should be noted that the stimuli design and labeling was further validated using participant responses in an iterative process (see \cite{saxe}). It is also noteworthy that less variance with GPT-4's ratings is anticipated as it is essentially a single rater with some random noise. What is remarkable is the degree of classification power achieved by GPT-4's appraisal ratings. 

To further examine the difference between GPT-4 and humans in perceiving the intended stimuli manipulation, we conducted a K-means cluster analysis using human ratings of the ten appraisal features. This way, we only rely on the model and participants' appraisal ratings to differentiate between classes of stimuli in a fully unsupervised manner. The analysis resulted in 12 and 20 stimuli clusters with Silhouette scores of 0.121 and 0.204 using human and GPT-4 ratings, respectively. This indicates that the reduced appraisal space is not enough to distinguish 20 classes (as intended by researchers) reliably based on human ratings; rather, they lead to 12 distinctive and fine-grained emotion labels. On the other hand, GPT-4 seems to exploit a latent learned appraisal mechanism that is able to decode the crafted stimuli into a fine-grained emotion space. Interestingly, ten appraisal features are enough to decode 20 clusters compared to the 12 using human ratings, even though the bottom-up clusters do not necessarily match the top-down/intended 20 emotion classes. Fig.~\ref{tsne_comp} illustrates the t-SNE plot generated using human and GPT-4 appraisal ratings (labeling is based on the majority intended emotion class making up each cluster). This plot illustrates the separability performance of GPT-4's appraisal features compared to humans, even in a two-dimensional space. This lends weight to the idea that GPT-4 is more consistent with the assumptions of appraisal theory than the individual human annotators.

\begin{table}[bt]
\caption{Correlation with human-reported appraisals}
\centering
\begin{tabular}{|l|c|l|c|}
\hline
\textbf{Feature} & \textbf{Value} & \textbf{Feature} & \textbf{Value} \\
\hline
Expectedness & 0.880\textsuperscript{***} & Agent Cause & 0.758\textsuperscript{***} \\
Pleasantness & 0.945\textsuperscript{***} & Self Cause & 0.887\textsuperscript{***} \\
Close Others & 0.904\textsuperscript{***} & Already Occurred & 0.528\textsuperscript{***} \\
Self-esteem & 0.912\textsuperscript{***} & Pressure & 0.892\textsuperscript{***} \\
Consequences & 0.839\textsuperscript{***} & Safety & 0.840\textsuperscript{***} \\
\hline
\multicolumn{4}{l}{$^{\mathrm{***}}p < .001$} \\
\end{tabular}
\label{app_corr}
\vspace{-10pt}
\end{table}

\begin{figure}[bt]
\centerline{\includegraphics[width=0.995\columnwidth]{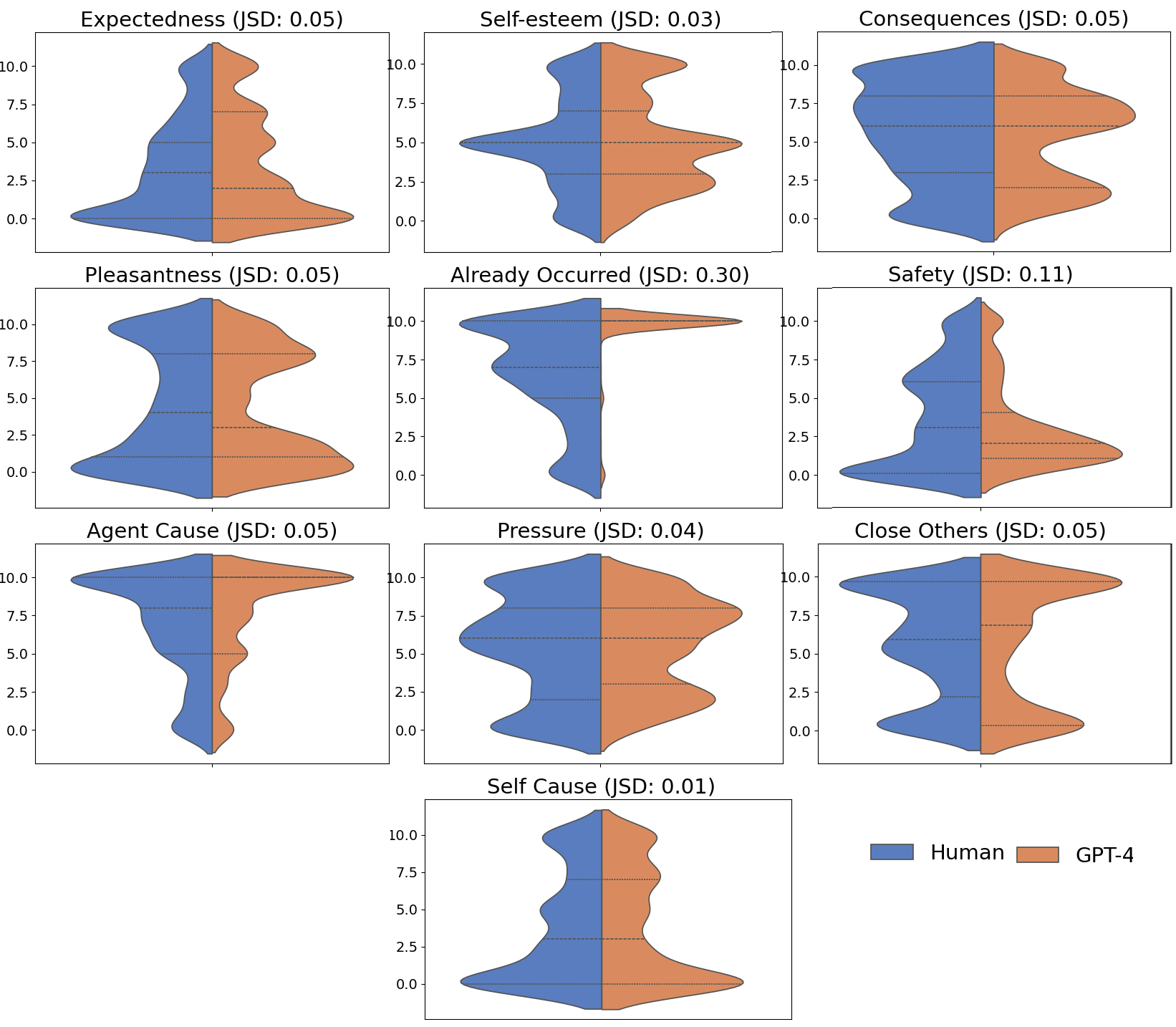}}
\caption{GPT-4 vs. human appraisal score probability distribution}
\label{viol_app}
\vspace{-10pt}
\end{figure}

\subsection{Appraisal derivation}
\begin{figure}[b]
\vspace{-10pt}
\centerline{\includegraphics[width=1.0\columnwidth]{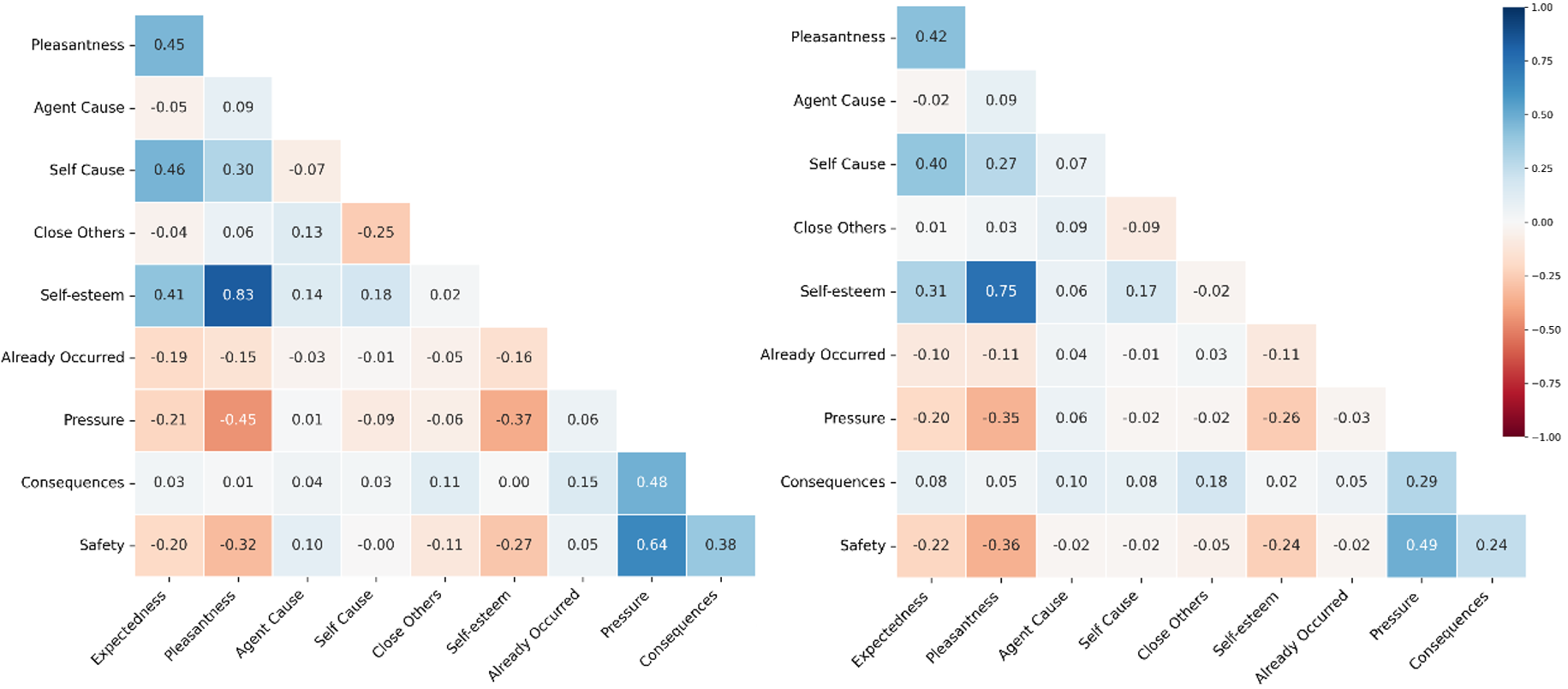}}
\caption{Inter-correlations of appraisals rated by humans, GPT-4, and the difference}
\label{app_inter_corr}
\vspace{-10pt}
\end{figure}

To examine how well GPT-4 predicts how a person would appraise a situation (i.e., appraisal derivation), we compute Pearson correlations between the ten appraisal variables rated by human participants and the corresponding variables predicted by GPT-4. To this end, we first averaged the scores over each stimulus for humans and GPT-4 to have a mean stimuli score. We observe very high correlations across the ten variables, suggesting the GPT-4 mean responses closely match human mean appraisal scores (Table \ref{app_corr}). GPT-4 seems to struggle to predict if a situation has already occurred (using the exact question formulation originally given to humans). To investigate the discrepancy further, we compared the rating distribution of all variables. Fig.~\ref{viol_app} shows the smoothed rating probability distribution with Jensen-Shannon Divergence (JSD) distance scores. We observe a higher concentration of GPT-4 ratings on either (or both) end of the spectrum, indicating higher variability among human raters with a relatively more tendency toward the mean. The density of already-occurred responses suggests significant disagreement among human participants over the event's time window. Given the remarkable degree of correspondence, this is a critical distinction between the two, warranting further investigation of the potential causes. We examine appraisal inter-correlations for further evidence of correspondence or deviation of behavior. However, Fig.~\ref{app_inter_corr} denotes impressive similarity between GPT-4 and human appraisal behavior with slight differences in pressure-safety and pressure-consequences inter-correlations. 

\subsection{Basic emotion recognition}\label{emo_recog}
\begin{figure}[b]
\vspace{-10pt}
\centerline{\includegraphics[width=0.995\columnwidth]{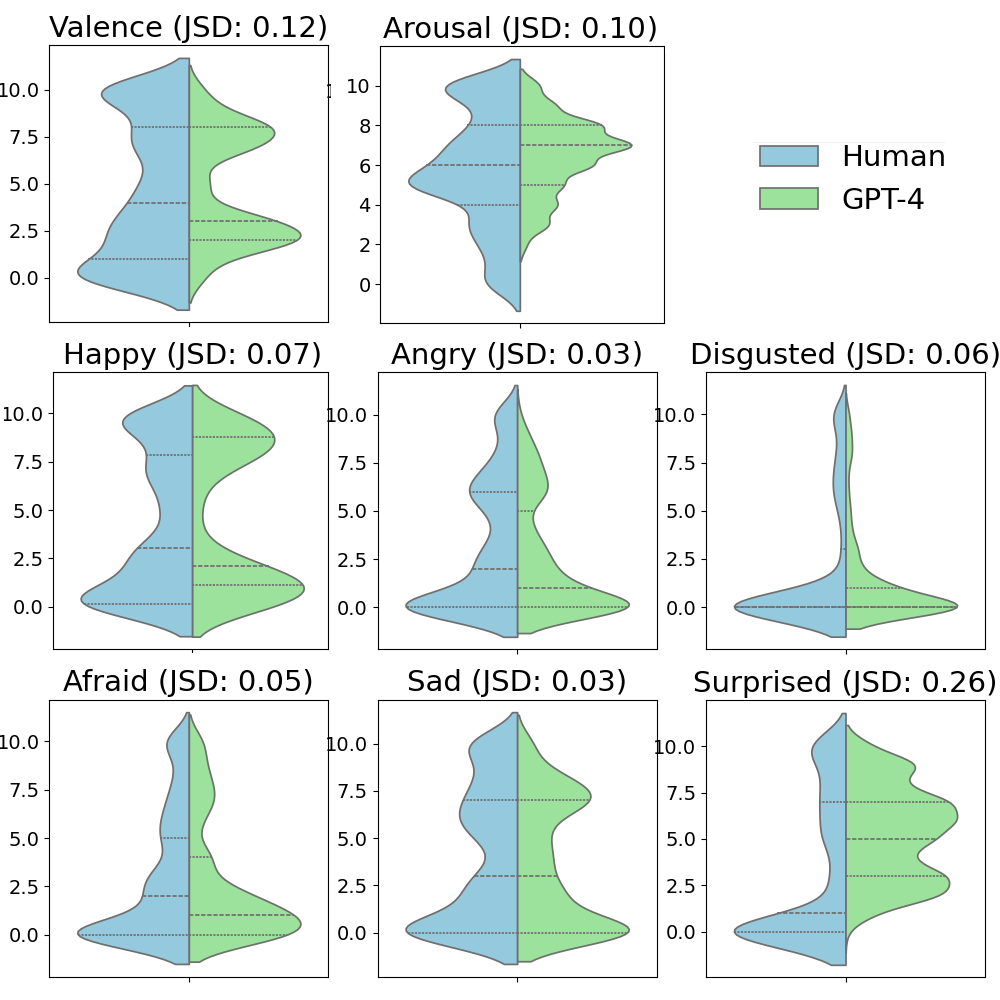}}
\caption{GPT-4 vs. human emotion score probability distribution}
\label{viol_emo}
\vspace{-10pt}
\end{figure}

Both participants and GPT-4 rated the stimuli on the eight basic emotion dimensions. Table \ref{emo_corr} demonstrates the results of Pearson correlation analysis. Similar to the appraisal derivation step findings, very significant correspondence is observed. Surprise scores the lowest among all emotion dimensions. Fig.~\ref{viol_emo} illustrates the smoothed probability distribution of emotion ratings and JSD scores. Corresponding to earlier observations on appraisal variables, we see a relatively greater tendency toward the mean among human participants except for surprise and, to a lesser extent, arousal. We observe a peek at higher arousal scores among human participants, which does not exist in GPT-4 distribution. GPT-4 surprise ratings denote a completely distinctive behavior compared to humans. GPT-4 tends to attribute significantly higher surprise to the stimuli. Such deviation requires deeper examination when combined with appraisals of the stimuli.

\begin{table}[bt]
\vspace{-10pt}
\caption{Correlation with human-reported emotion dimensions}
\centering
\begin{tabular}{|l|c|l|c|}
\hline
\textbf{Feature} & \textbf{Value} & \textbf{Feature} & \textbf{Value} \\
\hline
Valence & 0.963\textsuperscript{***} & Arousal & 0.846\textsuperscript{***} \\ 
\hline
Afraid & 0.930\textsuperscript{***} & Happy & 0.970\textsuperscript{***} \\
Angry & 0.887\textsuperscript{***} & Sad & 0.942\textsuperscript{***} \\
Disgusted & 0.885\textsuperscript{***} & Surprise & 0.671\textsuperscript{***} \\
\hline
\multicolumn{4}{l}{$^{\mathrm{***}}p < .001$} \\
\end{tabular}
\label{emo_corr}
\vspace{-10pt}
\end{table}

\subsection{Appraisal to emotion mapping}
\begin{figure}[bt]
\vspace{-10pt}
\centerline{\includegraphics[width=0.995\columnwidth]{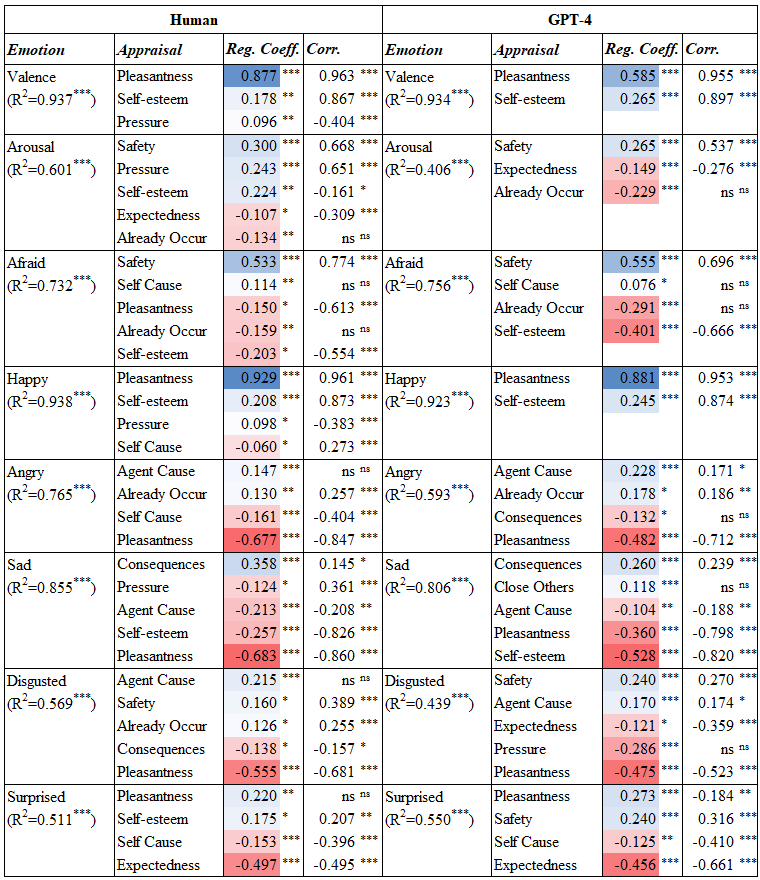}}
\caption{Human vs. GPT-4 tabular lens model}
\label{lens}
\vspace{-10pt}
\end{figure}

Finally, we investigate if GPT-4 reports a theoretically plausible relationship between appraisal variables and emotions. Recall that appraisal theories state that emotions arise from specific patterns of appraisals. Here, we examine and compare the pattern underlying human participants and GPT-4 responses. To this end, we conducted multiple linear regression (with backward elimination) to see if/how appraisals predict emotion dimensions. Additionally, for instances where regression coefficients might not offer clear insights, we supplemented our analysis with Pearson correlation to provide a more nuanced understanding of the relationships between variables, particularly in terms of shared variance. Results comparing GPT-4 to human responses are summarized in Fig \ref{lens}. Fig \ref{lens} is the tabular version of a lens model \cite{lens}, which shows how well each emotion dimension is predicted from the appraisal ratings, what appraisal variables contribute the most in predicting each emotion dimension (top five, if statistically significant), and the correlation of the top predictors and each emotion dimension. 

We see a remarkable match between the overall mapping behavior, which is consistent with the observation reported in \cite{ala1,ala2}. The variance of all emotion dimensions is explained to a great extent by the reduced appraisal space. Surprise has the least explained variance among human emotions, whereas arousal and disgust require a larger appraisal space to be fully explained by GPT-4 ratings. Regarding major appraisal-emotion mapping differences, pressure, and self-esteem seem to be important predictors of arousal with respect to human ratings but are not significant based on GPT-4 ratings. Similarly, pressure and self-cause are predictors of the human happy dimension but are not predictors of GPT-4 happy scores. GPT-4 leverages its evaluation of close others' involvement in stimuli to predict sadness, a role that pressure plays when it comes to human attribution of sadness to stimuli. GPT-4 attributes disgust to cases that are not expected, whereas human participants highlight the role of long-term consequences when assigning disgust to stimuli. It should be noted that the stated differences are negligible when compared to notable affect derivation pattern similarities between the GPT-4 and human participants. 

\subsection{Discussion}
In Study 1, we employed a dataset of crafted stimuli rated by external observers, which were systematically manipulated in ways that yield different appraisals and emotions. Except for a few instances, results suggest a remarkable correspondence of mean human scores and mean GPT-4 scores in all processes involved in emotional cognition. The similarities exceed what is reported in \cite{ala1,ala2} and others. One interpretation might be that GPT-4 excels when dealing with stereotypical situations rather than free-form self-report idiosyncratic vignettes. Also, GPT-4 might view situations as an observer and capture the third-person perspective of the average human.

\section{Study 2: Investigating the perspective}
Study 2  tests the hypothesis,  suggested by Study 1, that GPT-4 processes emotions through the lens of an average observer. To this end, we employ the crowd-enVENT dataset developed by Troiano et al. \cite{enVent1}, which is, to the best of our knowledge, the only corpus that includes both author and reader annotations of both appraisals and emotions. This valuable corpus enables researchers to compare the agreement of external annotators and self-assessments of the authors. Similar to the first study's corpus, the appraisal scheme used to create crowd-enVENT is primarily based on the scheme proposed by Scherer and colleagues \cite{scherer2,scherer3}. Unlike the first study, crowd-enVENT consists of self-reported vignettes (1200 data points). However, strategies are adopted to promote the collection of more idiosyncratic events to induce a higher diversity of events and appraisal dimensions  \cite{enVent1}. In this corpus, participants rate 21 appraisal variables on a scale of 1-5 and pick an emotion from 12 emotion labels plus a ``no emotion" label. For the sake of brevity, we refer readers to the original paper for detailed descriptive statistics and corpus creation and validation processes. Here, we focus on testing our hypothesis that GPT-4 is aligned more with an average observer's evaluation of emotions and appraisal induced in events.

\subsection{Appraisal derivation}\label{AA}
\begin{figure}[tb]
\vspace{-10pt}
\centerline{\includegraphics[width=1.0\columnwidth]{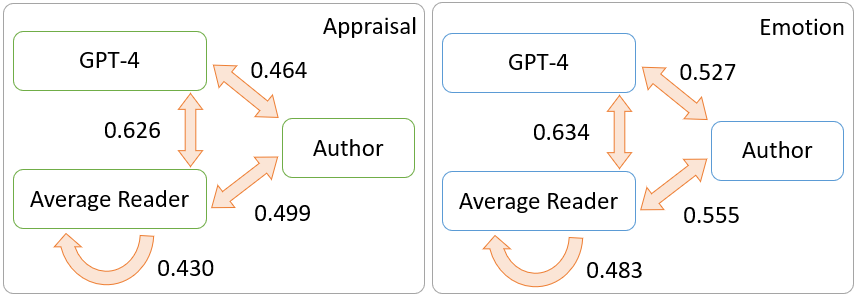}}
\caption{Comparison of emotion classification and appraisal rating agreement (Krippendorff's alpha)}
\label{enVent_emo}
\vspace{-10pt}
\end{figure}

We first turn our attention to the correspondence of appraisal ratings between GPT-4, the author's self-assessments, and external human raters. We follow the same procedure as described in Study 1 to generate prompts with minimum additional text for standardized output. Fig \ref{enVent_emo} provides a comparison of appraisal rating agreements using Krippendorff's alpha measure \cite{kripp}. Krippendorff's alpha is particularly suitable here due to its ability to handle different levels of measurement (nominal, ordinal, interval, and ratio and to accommodate any number of raters. It also enables to have the same agreement measure for both appraisals and emotions. Based on Fig \ref{enVent_emo}, we see GPT-4 corresponds significantly better to the average ratings of readers than the authors' original self-assessments. The average reader (considered a single rating) slightly outperforms GPT-4. However, we see the lowest agreement levels among readers (5-way), suggesting high variations in evaluations among human external observers of an emotion-evoking event. In summary, GPT-4 seems to be on par with an average human observer and significantly more in line with a third-person perspective than the self-assessment of the event.

\begin{figure}[tb]
\centerline{\includegraphics[width=0.9\columnwidth]{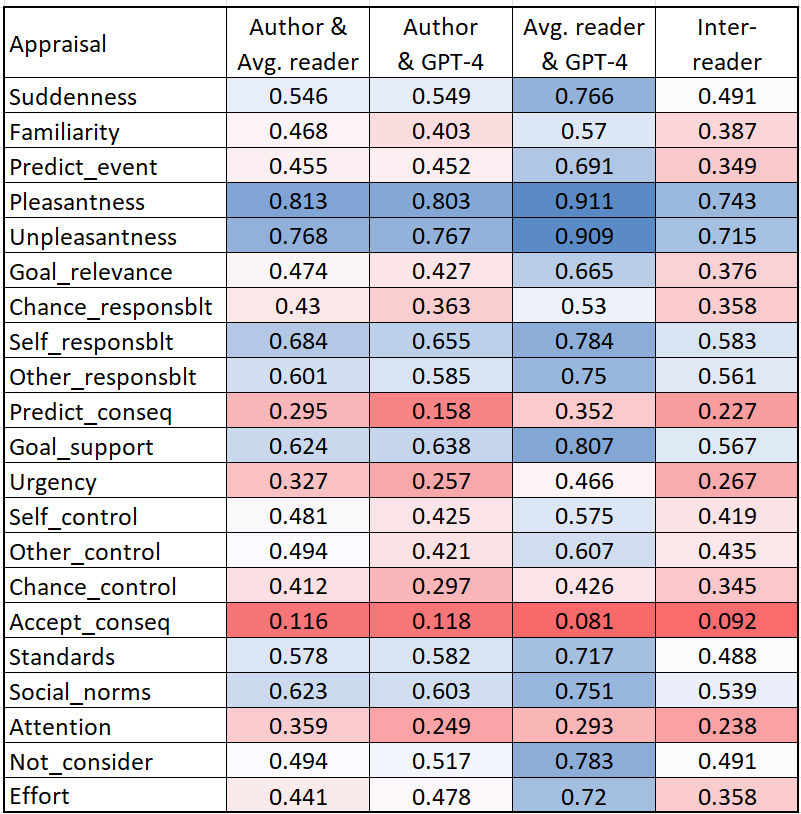}}
\caption{Tabular comparison of agreement across appraisal variables (Krippendorff's alpha)}
\label{envent_app}
\vspace{-10pt}
\end{figure}

To delve into the nuances, Fig. \ref{envent_app} showcases the agreement scores for the 21 appraisal variables. GPT-4 aligns more closely with the average third-person perspective in 20 of the 21 appraisals when comparing the "Avg reader \& GPT-4" and "Author \& GPT-4" columns. GPT-4 mostly matches the average reader's predictions of the author's self-assessed appraisals or exceeds that in 6 out of the 21 appraisal variables. The lower inter-reader agreements on appraisal features emphasizes individual differences and indicates GPT-4's tendency towards a balanced average-human assessment of emotional situations. Furthermore, a consistent pattern across the four columns indicates that certain appraisals are universally challenging to predict, whether from an individual or a generalized social perspective, hinting at the need for additional situational or personal information for precise appraisal prediction. Variables such as pleasantness, unpleasantness, goal-support, and congruence, as well as self/other responsibility, are relatively straightforward and show higher agreement levels across comparisons. Conversely, appraisal variables like accept\_consequence (i.e., accommodative coping) are more dependent on the individual involved. These subtleties call for further research to enhance LLMs' emotional cognition by incorporating individual variability.

\subsection{Emotion recognition}\label{FT}
The authors, five human readers, and GPT-4 attributed a label among 13 emotion labels to the self-reported vignettes. Similar to the previous step, we employ Krippendorff's Alpha in this classification task, as it calculates the degree of agreement corrected for the chance to ensure that the agreement among raters is not simply due to random chance. We employ the most representative label for a given vignette (i.e., majority rule) to serve as a proxy for the collective attribution of labels to scenarios or the consensus viewpoint. Fig \ref{enVent_emo} demonstrates the results of the analysis. Similar to the appraisal derivation step, GPT-4 corresponds significantly more to the majority class selected by readers than the authors' original self-assessments and is on par with average readers in predicting original labels (relative to inter-reader agreement). 






\section{General Discussion}
In the two studies conducted, we explored how GPT-4 processes and interprets emotions from both artificially crafted and self-reported textual narratives following the method presented in \cite{ala1,ala2}. Our approach illuminates the mechanisms by which how humans and LLM differ in terms of appraising situations, attributing emotions to situations, and, lastly, mapping appraisals to emotions. Tak and Gratch \cite{ala1, ala2} applied this approach to autobiographical memories, focusing on how well the two models predict self-reported appraisals and emotions. The results showed the model did not fully recover the significance of all appraisal dimensions and relied only on a few. They also reported that the models were less accurate at predicting arousal than valence and dominance dimensions. However, in \cite{ala1,ala2}, participants were instructed to recall emotional situations, whereas GPT was given no information about the nature of these situations, including the event's significance to the person. In addition, participants self-selected their own stories, which could inflict biases in what type of stories are remembered. For example, the stories might not vary widely in certain appraisal dimensions.
To address these limitations, in Study 1, we employed 200 stimuli describing situations that evoke particular emotions, initially developed to understand brain representations of emotions. GPT-4 was prompted to rate these scenarios on appraisal and emotion scales used by human participants. A feature reduction was performed to identify key appraisal dimensions influencing emotion recognition.
Remarkably, GPT-4 demonstrated high accuracy (99.7\% with full appraisal space) in classifying emotions, surpassing human participants. Correlation analysis revealed GPT-4's ratings closely aligned with human appraisals across ten variables, though differences in interpreting the temporal aspect of events were noted (i.e., whether an event has already occurred). Differences in how GPT-4 and humans evaluate certain emotions and appraisals were highlighted, especially in the context of surprise and arousal. Given the remarkable correspondence across appraisals and emotions, such discrepancies warrant further investigation. In the second study, focusing on whether GPT-4 aligns more with an average observer's evaluation, we utilized another corpus featuring both author and external reader annotations of emotions and appraisals. GPT-4's appraisals were more aligned with average readers than authors’ self-assessments, indicating a closer association with an observer’s perspective. In emotion recognition, GPT-4 again showed closer alignment with average readers’ viewpoints compared to authors' self-assessments.

Most recent evaluations of LLMs' emotion-related abilities generally report higher-than-expected performance. Yongsatianchot et al. \cite{nut1} reported GPT-4 can generate diverse emotional stories, explain emotional events, and do reverse appraisal when given enough context. Broekens et al. \cite{broekens} reported ChatGPT zero-shot performance matches fine-tuned XLMRoBERTa-large in VAD prediction, and it can generate new valid situations based on levels of latent affect representations. According to Elyoseph et al. \cite{elyoseph}, ChatGPT scores significantly higher than those of the general population on the Levels of Emotional Awareness scale. Nevertheless, some limitations are also raised. Several studies highlighted the sensitivity of the LLMs' performance to variations in the prompt formulation \cite{ala1}\cite{broekens}. LLMs are also found to be biased or limited in various aspects. GPT models found to rate the negative valence to be more negative than humans \cite{nut2}. GPT-4 struggled to predict relevance, emotion intensity, and coping responses self-reported by humans and struggled to reason correctly about how emotion has consequences for decision-making \cite{ala1}.

The mentioned studies have either used corpora with self-annotated emotions or external observers' annotation as the ground truth. From the first-person perspective, we are bound to rely on participants' memories to recall their feelings from past experiences or their self-assessments in a controlled situation. However, participants often recall intense events or long-lasting emotions and might be unable to report their feelings accurately as they happened in the distant past or contain sequential and varied emotions. The latter approach is also challenging due to the difficulty and the cost of running studies to evoke emotions in controlled settings.

In the third-person perspective, annotators are expected to predict the emotions of others with the limited available context. Hence, the ground truth represents how participants predict the emotions of others rather than the true experienced emotion. We showed that GPT-4 has essentially learned this general observer's viewpoint. Earlier work has also reported that LLMs are better at approximating ``average human judgments" \cite{dillion2023can, santurkar}, the ``wisdom of small crowds" \cite{trott}, or the ``aggregate summary of human knowledge"\cite{ong} than they are at capturing variation and human diversity \cite{abdurahman}. Although, at first glance, the third-person perspective might appear as a silver standard, it can offer a level of objectivity and standardization that is challenging to achieve with self-reported data with inherent biases and personal contexts. In the absence of additional individual information and computational models that can reliably account for such individual variability, the third-person perspective is aptly attuned to a more general emotional experience. Furthermore, using third-person data might potentially pose fewer ethical concerns and privacy issues compared to self-reported data, which can be more sensitive and personal. On the other hand, this approach could lead to over-generalization, potentially resulting in reduced diversity and inclusion, and may also perpetuate biases in social perceptions.



\section*{Ethical Impact Statement}
This paper re-analyzed previously collected de-identified data previously subjected to ethical review. This data is used as a benchmark to scrutinize the underlying mechanisms of how pre-trained language models process human emotion. However, it should be noted that we investigated a single language model, which is constantly being updated; hence, caution must be taken in generalizing these findings to other language models or other versions of the examined model. As predicted by prior research on emotion, strong cultural and demographic differences exist in how emotional situations are construed. Thus, these findings should be replicated across these different groups. Finally, the findings highlight the potential concerns for those seeking to deploy large language models to reason about human emotion or generate emotional content. Given the criticality of potential harm caused by the emotional manipulation of LLMs (or any AI, for that matter), we need constant measurements of LLMs' emotional cognition and manipulation abilities with every new LLM or any updates to current LLMs. We need to have a comprehensive benchmark for such studies.

\section*{Acknowledgment}
This work is supported by the Army Research Office under Cooperative Agreement Number W911NF-20-2-0053. The views and conclusions contained in this document are those of the authors and should not be interpreted as representing the official policies, either expressed or implied, of the Army Research Office or the U.S. Government. The U.S. Government is authorized to reproduce and distribute reprints for Government purposes notwithstanding any copyright notation herein.

\vspace{12pt}

\end{document}